**A VLM-Enhanced Framework for Comprehensive Traffic Sign Condition Assessment Integrating Daytime Visual Performance and Nighttime Retroreflectivity Evaluation**

**Linlin Zhang**
Post Doctoral Fellow
Department of Civil and Environmental Engineering
University of Missouri- Columbia, Columbia, MO, USA, 65201
Email: linlinzhang@missouri.edu

**Neema Jakisa Owor**
Ph.D. Candidate
Department of Civil and Environmental Engineering
University of Missouri- Columbia, Columbia, MO, USA, 65201
Email: nodyv@missouri.edu

**Xiang Yu**
Ph.D. Student
Department of Civil and Environmental Engineering
University of Missouri- Columbia, Columbia, MO, USA, 65201
Email: xytm4@missouri.edu

**Abby Watts**
Master Student
Department of Civil and Environmental Engineering
University of Missouri- Columbia, Columbia, MO, USA, 65201
Email: arw5dv@missouri.edu

**Yaw Adu-Gyamfi**
Associate Professor
Department of Civil and Environmental Engineering
University of Missouri- Columbia, Columbia, MO, USA, 65201
Email: adugyamfiy@missouri.edu

## ABSTRACT

Traffic signs are crucial components of road safety, serving as visual tools under all lighting conditions. The Manual on Uniform Traffic Control Devices (MUTCD) specifies daytime visual factors such as legibility and color contrast, and nighttime retroreflectivity requirements. Traditional assessment methods rely on manual inspections, which the Federal Highway Administration (FHWA) notes are subjective, labor-intensive and pose safety concerns, while retroreflectometers are expensive and unaffordable for smaller agencies. Most existing studies focus on either daytime factors or nighttime retroreflectivity but rarely integrate both aspects comprehensively. This study develops a novel framework that systematically evaluates traffic signs through integrated daytime-nighttime assessment. The methodology employs three fine-tuned Vision Language Models (VLMs) for daytime visual performance assessment across four key factors: legibility, color, surface and shape integrity, and surrounding environment conditions. VLM predictions are converted to numerical scores through sentiment analysis and Contrastive Language-Image Pre-Training (CLIP) scoring, while nighttime performance is assessed using LiDAR-derived retroreflectivity following established calibration procedures. The framework integrates these components into a comprehensive Sign Condition Index (SCI) for maintenance guidance. Evaluation results demonstrated that LLaVA and Qwen outperformed InternVL, achieving bidirectional cosine similarity scores of 0.67-0.76 across all factors. Among 462 validated traffic signs, 68 were flagged by the proposed framework as requiring immediate replacement due to inadequate retroreflectivity performance. This research provides a cost-effective alternative to traditional manual inspections for comprehensive traffic sign condition assessment.

**Keywords:** Traffic Sign Assessment, Vision Language Models, Retroreflectivity, MUTCD Compliance, Transportation Infrastructure Management, LiDAR Retro-Intensity

## INTRODUCTION

Cities across the United States maintain extensive inventories of traffic signs, with evaluations for Manual on Uniform Traffic Control Devices (MUTCD) (*1*) compliance being resource-intensive and costly, particularly straining smaller municipal budgets. Traffic signs are crucial components of road safety and traffic management, serving as visual tools to guide and regulate drivers, pedestrians and cyclists under all lighting conditions. During the day, clear legibility, vivid color contrast, intact surface and shape integrity and an unobstructed surrounding environment ensure that text and symbols can be read quickly and accurately understood by all road users. Each of these four daytime factors plays a vital role in sign conspicuity and driver comprehension. At night or in adverse weather, sufficient retroreflectivity becomes equally important, as it reflects headlight illumination to drivers's eyes, thus enabling prompt detection of traffic signs. Environmental exposure and physical damage often degrade traffic signs, necessitating regular inspections to uphold these standards. Color fading and loss of retroreflectivity generally occur together as concurrent mechanisms of traffic sign degradation (*2*). The MUTCD provides comprehensive specifications for these performance requirements, including minimum letter heights for legibility, chromaticity coordinate limits for color contrast, provisions for surface and shape integrity, obstruction clearance requirements for environmental conditions, and minimum retroreflectivity thresholds for nighttime visibility, all designed to promote consistent visibility and compliance across all lighting conditions.

Traditional methods of traffic sign condition assessment have heavily relied on manual inspection procedures, where inspectors physically examine signs for wear, damage, or obstruction. Specifically, retroreflectivity assessment typically involves either nighttime visual inspections from moving vehicles or handheld retroreflectometers. However, the Federal Highway Administration (FHWA) has explicitly noted that visual inspections are highly subjective labor-intensive and pose safety concerns (*3*). Meanwhile, retroreflectometers, though accurate, are expensive and often unaffordable for smaller agencies. For smaller municipalities, such resource requirements pose significant budgetary challenges, as highlighted by the Minnesota Local Road Research Board (LRRB) and FHWA (*4*, *5*). Moreover, comprehensive assessment of the conditions of signs presents significant operational challenges, as the diverse nature of daytime evaluation factors requires different assessment techniques and expertise, while nighttime retroreflectivity evaluation demands specialized equipment and standardized procedures that differ substantially from visual inspection methods. This complexity necessitates the development of integrated assessment approaches that can address both daytime and nighttime evaluation requirements efficiently.

Although recent technological advances have facilitated the development of automated approaches for traffic sign analysis, research on comprehensive condition assessment remains limited. Most existing studies adopted computer vision and deep learning techniques but focus mainly on daytime factors, such as legibility and color. Specialized research has emerged on retroreflectivity prediction, with studies employing artificial neural networks and deep neural networks to model the degradation of retroreflectivity over time (*6*, *7*). Moreover, integrated approaches combining LiDAR and digital image have also been explored. For example, (*8*) implemented this approach for traffic sign detection and classification instead of condition assessment. (*9*) employed the combined image and LiDAR data for automated retroreflectivity evaluation and established categorical assessment protocols. The camera images were used primarily for color segmentation to associate LiDAR points with different sign colors rather than conducting visual condition analysis. This method focused solely on retroreflectivity assessment without evaluating daytime visual performance factors such as legibility, shape integrity, or environmental conditions. Overall, these above studies addressed isolated aspects, either specific factors or sign detection and classification, a comprehensive understanding of the overall state of traffic signs across both daytime and nighttime is still lacking.

To address these limitations, this study develops a comprehensive framework that systematically evaluates traffic signs through integrated daytime-nighttime assessment. We employ fine-tuned Vision Language Models (VLMs) for daytime visual performance assessment across four key factors: legibility, color, surface and shape integrity, and surrounding environment conditions. VLM predictions are converted into numerical scores through sentiment analysis and Contrastive Language-Image Pre-Training (CLIP) (*10*) scoring, while nighttime performance is assessed using LiDAR derived retroreflectivity that is normalized and converted to estimated retroreflectivity following Ai and Tsai (*11*). The effectiveness of the fine-tuned VLMs is evaluated by cosine similarity score. This framework provides standardized and objective evaluations, reducing the need for manual inspections and costly specialized equipment while offering practical guidance for traffic sign maintenance decisions.

The contributions of this study are significant and represent important advancements in the field of traffic sign condition assessment, particularly in the context of comprehensive daytime and nighttime evaluation. This study makes substantial contributions in three key areas as follows:

1. We introduce a comprehensive framework that integrates daytime visual performance assessment with nighttime retroreflectivity evaluation, addressing MUTCD related factors within a unified system rather than examining individual components in isolation. This framework systematically combines multiple assessment dimensions that have traditionally been evaluated separately, achieving complete traffic sign condition analysis.
2. We develop a novel methodology that employs fine-tuned VLMs for daytime visual assessment across four critical factors, coupled with LiDAR based retroreflectivity estimation for nighttime evaluation. The approach incorporates a hybrid scoring system that converts descriptive VLM predictions into standardized numerical scores through sentiment analysis and CLIP scoring, while utilizing LiDAR retro-intensity data converted to estimated retroreflectivity following established calibration and estimation methods.
3. The proposed framework provides a cost-effective, scalable and more standardized solution that significantly reduces dependence on manual inspections and expensive specialized equipment, thereby offering municipal agencies an accessible and standardized tool for traffic sign maintenance prioritization and decision-making.

The remainder of this paper is structured as follows. First, an overview of related works. The next section presents the methodology of this work. Then, evaluation and results are discussed. The final section concludes findings, remarks, and future research directions.

## LITERATURE REVIEW

Although the MUTCD provides detailed specifications for the assessment of traffic signs, the evaluation process remains labor-intensive, time-consuming, and subjective. To reduce costs, save time, and enhance evaluation consistency, many studies have explored robustness assessment methods. Existing research on traffic sign robustness assessment can be broadly divided into two categories: physical damage and retroreflectivity assessment, and large language models (LLMs) and VLMs applications in transportation.

### Physical Damage & Retroreflectivity Assessment

Several studies have explored automated methods for traffic sign condition assessment. (*12*) integrated mobile laser scanning (MLS), deep learning techniques, and MLS point clouds, to identify physical damage on traffic signs. This approach effectively identified signs with physical issues such as tilted poles, deformed boards, and signs that had fallen. However, it did not address more subtle forms of deterioration, such as reflective wear or gradual weathering of the sign material. Additionally, the system's reliance on MLS data made it potentially too resource-intensive for routine monitoring. To address it, (*13*) leveraged computer vision approach to assess signs exposed to various forms of physical damage, including rust, graffiti, and cracks. The approach had two steps, You Only Look Once version 8 (YOLOv8) for detection and a convolutional neural network (CNN) for damage classification. However, challenges like dataset imbalance and lack of retroreflectivity assessment remain. (*11*) developed an automated method for evaluating traffic sign retroreflectivity by using a mobile LIDAR system combined with computer vision techniques. This approach significantly enhanced data collection efficiency compared to traditional manual inspections. But despite its advantages, the model was sensitive to environmental conditions. Building on these advancements, (*14*) used artificial neural networks (ANNs) to improve the detection of subtle issues like rust and cracks on signs. Retroreflectivity assessment has also advanced. But challenges still remain such as the daytime factors and adapting to environmental conditions.

### Large Language and Vision Language Models in Transportation

Recent advances in large LLMs and VLMs have expanded their applications in transportation tasks involving reasoning, prediction, perception, and decision support. Existing studies have primarily explored LLMs for traffic flow forecasting and traffic signal control, while the use of VLMs for image-based transportation infrastructure assessment remains comparatively limited. In recent years, although the deep learning and neural network models are becoming increasingly complicated with more multi-model traffic data encoded, improvements in traffic flow prediction accuracy remains limited (*15–19*). To enhance traffic flow prediction explainability, (*15*) proposed xTP-LLM, converting multi-model traffic data into neural language description. (*16*)explored a different perspective by leveraging LLMs to improve the adaptability of traffic prediction models under varying traffic conditions. Large Language Model Enhanced Traffic Flow Predictor (LEAF) model was introduced, using a frozen LLM as a selector in a dual-branch graph framework to improve adaptability. (*17*) further explored LLMs as generators by encoding graph features into prompts for direct prediction. Building on this, (*18*) developed Spatial-Temporal LLM (ST-LLM) with spatial-temporal tokenization and partially frozen attention. (*19*) extended it to ST-LLM+ by incorporating graph attention (PFGA) and LoRA-based training, achieving state-of-the-art results on real-world datasets. Recent studies have explored the integration of LLMs into traffic signal control systems from various perspectives. (*20*) proposed a framework based on LightGPT, which learns traffic patterns and control strategies. The LLM is guided by a knowledgeable prompt that incorporates real-time traffic conditions.(*21*) introduced an LLM-based model for Adaptive Traffic Control Systems (ATCS), enhancing reasoning and planning using knowledge from dynamic intersections. (*22*) developed a hybrid framework that combines LLMs with perception and decision-making tools for human-like control in complex urban settings. (*23*) applied an advanced LLM to implement green wave coordination on urban arterials, leveraging its reasoning ability to optimize signal timing.

However, current research exhibits notable limitations. Most studies focus on either daytime visual performance assessment using image-based methods or nighttime retroreflectivity evaluation but rarely integrate both aspects for comprehensive traffic sign condition assessment. While some approaches have integrated both devices, image-based data are used for detection or segmentation rather than evaluation purpose. Furthermore, although LLMs have demonstrated broad applications in traffic flow prediction and signal control, VLMs remain underexplored for image-based traffic sign condition assessment. To address these gaps, this study presents an integrated framework that combines daytime visual performance evaluation with nighttime retroreflectivity assessment, using fine-tuned VLMs to support daytime visual condition assessment and LiDAR-derived retroreflectivity for nighttime evaluation.

## METHODOLOGY

This study proposes a comprehensive framework for traffic sign condition assessment that integrates daytime visual performance evaluation with nighttime retroreflectivity assessment, as illustrated in **Figure 1**. The framework consists of three sequential components designed to provide standardized and objective traffic sign condition evaluation.

The first component involves a two-phase data collection strategy that supports both evaluation modes. Phase one focuses on gathering image data for VLM fine-tuning, while phase two involves simultaneous collection of image and LiDAR fusion data for comprehensive evaluation. The second component comprises the dual-mode evaluation process, which operates through two parallel approaches: daytime visual performance assessment using fine-tuned VLMs and nighttime retroreflectivity assessment using LiDAR-derived measurements. The third component establishes the SCI through hybrid scoring systems that convert evaluation results into numerical scores and integrate them into practical maintenance recommendations.

This integrated approach addresses the limitations of traditional manual inspection methods by providing cost-effective, objective, and comprehensive assessment capabilities that evaluate traffic signs under both daytime and nighttime conditions within a unified framework.

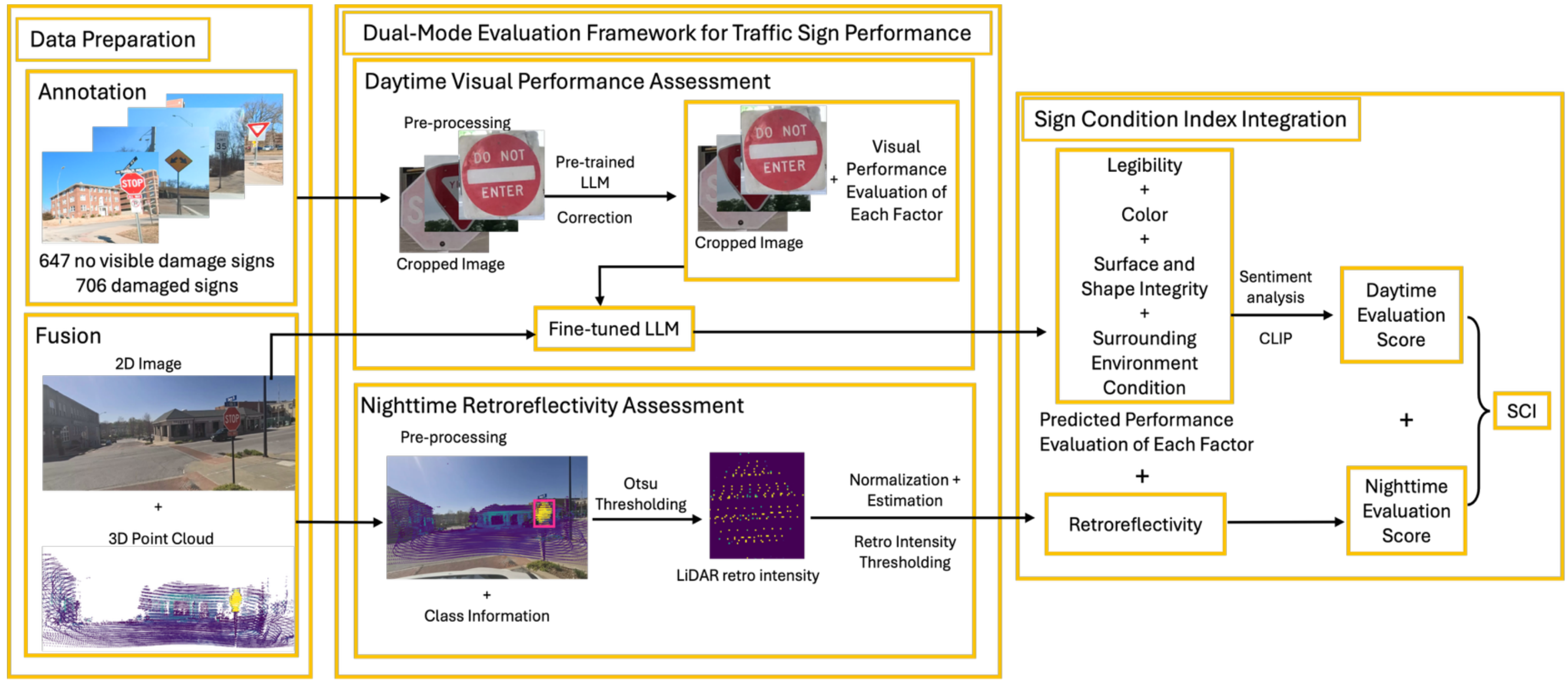

**Figure 1. Proposed Pipeline**

**Data Collection and Preparation**

*Two-Phase Data Collection Strategy*

Data collection was conducted in two distinct phases on urban roads throughout Columbia, Missouri, to support the comprehensive evaluation framework.

The first phase focused on gathering traffic sign image data for fine-tuning the VLMs. We captured various types of traffic signs under clear weather conditions and adequate lighting to ensure optimal image quality for daytime visual assessment training. A GoPro 11 Black camera was used in linear mode with a frame rate of 30 Hz to maintain consistent image acquisition parameters. This phase resulted in the collection of 647 signs showing no visible damage and 89 manually collected signs with various types and levels of damage, providing the foundation dataset for VLM fine-tuning.

The second phase involved simultaneous collection of both image and LiDAR data. Image acquisition continued using the same GoPro 11 Black camera configuration. A Livox HAP LiDAR was used to collect high-resolution point cloud data with retro-intensity value that suitable for nighttime retroreflectivity evaluation. The sensor was configured for synchronized acquisition with the camera. **Figure 2** shows the integrated LiDAR and camera setup used for multi-sensor data collection. To achieve effective data fusion between the two sensors, we utilized the Robot Operating System (ROS) for temporal synchronization.

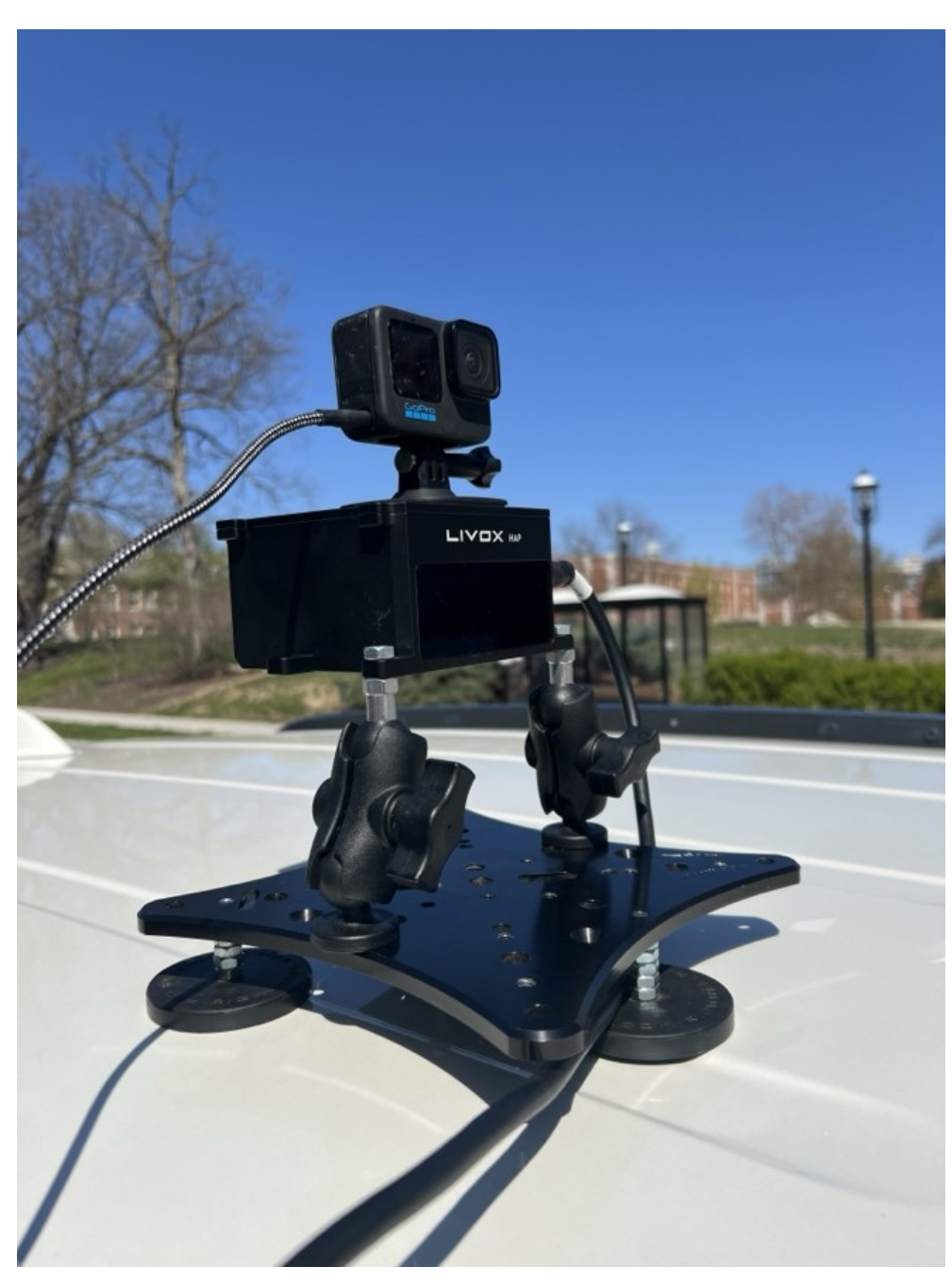

**Figure 2. LiDAR and Camera Setup**

*Dataset Preparation and Augmentation*

Since most manually collected traffic signs exhibited good condition, achieving a balanced dataset required additional data sources to ensure unbiased model performance. A balanced dataset comprising equal proportions of good and damaged signs is essential for robust model training.

To address the imbalance between good and damaged signs, we incorporated a public damaged traffic sign dataset (*24*) and applied systematic augmentation techniques to 300 images of signs in good condition. The public dataset contributed 317 damaged signs from various real-world scenarios.

The augmentation process targeted the four evaluation factors, with each factor divided into three severity levels: slight, moderate, and severe. Different combinations of these factors and severity levels generated various degradation effects including rain drops, hue shifts, motion blur, sun exposure, etc. Among the 300 augmented signs, 270 images had one randomly selected augmentation effect applied, 20 images had two randomly selected effects applied simultaneously, and 10 images had three randomly selected effects applied simultaneously. Each augmentation effect corresponds to a specific factor at a particular severity level, as detailed in **Table 1**.

While each augmentation effect is categorized under a specific factor in **Table 1**, many effects inherently impact multiple evaluation factors. For example, vegetation occlusion listed under Surrounding Environment Condition also affects Legibility by reducing text readability. During the annotation process, these cross-factor impacts were manually identified and corrected to ensure accurate labeling across all affected factors.

**Table 1. Augmentation Effects by Factor and Severity Level**

| **Factor** | **Severity Level** | **Effect** | **Visual Effect Description** |
|---|---|---|---|
| Legibility | Slight | Light Gaussian Blur | Mimics dust, aging, or mild glare |
| | | Lower Contrast | Mild visual fatigue or dull color from wear |
| | | Faint Transparent Sticker | Mild obstruction or sticker aging |
| | Moderate | Motion Blur | Simulates rainy day blur or camera shake |
| | | Text Occlusion | Dirt spots or partial occlusion of text |
| | Severe | Blur and Patch | Simulates heavy dirt or smeared glass |
| | | Text Cracking | Characters become unreadable, overall image still clear |
| | | Overexposed and Faded | Mimics sun-faded, aged signs |
| | | Graffiti | Strong effect of malicious defacement |
| Color | Slight | Mild Desaturation | Mimics natural fading or sun bleaching |
| | | Warm Filter | Faded red/yellow signs due to oxidation or age |
| | Moderate | Strong Desaturation | Loss of color clarity in poor lighting |
| | | Hue Shift | Color distortion due to sensor or weather |
| | | Contrast Drop | Color dullness in dirty or old signs |
| | Severe | Color Inversions | Severe chemical or reflective surface damage |
| | | Channel Dropout | Simulates sensor failure or extreme aging |
| | | Saturation Banding | Severe UV damage or paint stripping |
| Surface and Shape Integrity | Slight | Edge Wear | Slightly worn, dull, or softened edges |
| | | Slight Distortion | Mostly intact shape with minor irregularity |
| | | Minor Surface Dents | Sign surface not smooth but overall shape intact |
| | Moderate | Crooked Frame | Sign appears mounted at an angle |
| | | Warped Shape | Sign is clearly deformed, skewed or not planar |
| | | Edge Fragmentation | Border looks torn or fragmented at multiple points |
| | | Sticker | Partial content blocked by stickers |
| | Severe | Broken Corners | Corners missing, broken, or cracked |
| | | Extreme Deformation | Sign looks melted, or deeply misshapen |
| | | Structural Collapse | Contour collapse / segmentation loss |
| | | Bent | Extreme viewing angle |
| Surrounding Environment Condition | Slight | Light Shadow | shadowed by branches or poles |
| | | Light Reflection/Surface Glare | Bright reflections or glare |
| | Moderate | Branch Occlusion (Edge-based) | Branches partially covering sign edges or corners |
| | | Rain Droplets | Water droplets on sign surface reducing clarity |
| | Severe | Heavy Vegetation Occlusion | Dense foliage or overgrown vegetation significantly blocking sign visibility |
| | | Partial Blocking by Poles/Objects | Infrastructure elements or objects partially obstructing sign content |

The complete dataset for VLM fine-tuning included 647 undamaged sign images, 89 manually collected damaged signs, 317 damaged signs from the public dataset, and 300 augmented signs with degradation effects. In total, we obtained 706 damaged sign images and 647 undamaged sign images, providing the balanced dataset required for model training. **Figure 3** demonstrate the images from different data source. We split this dataset into training and testing sets using a 70%-30% ratio, with 941 samples for training and 412 samples for testing.

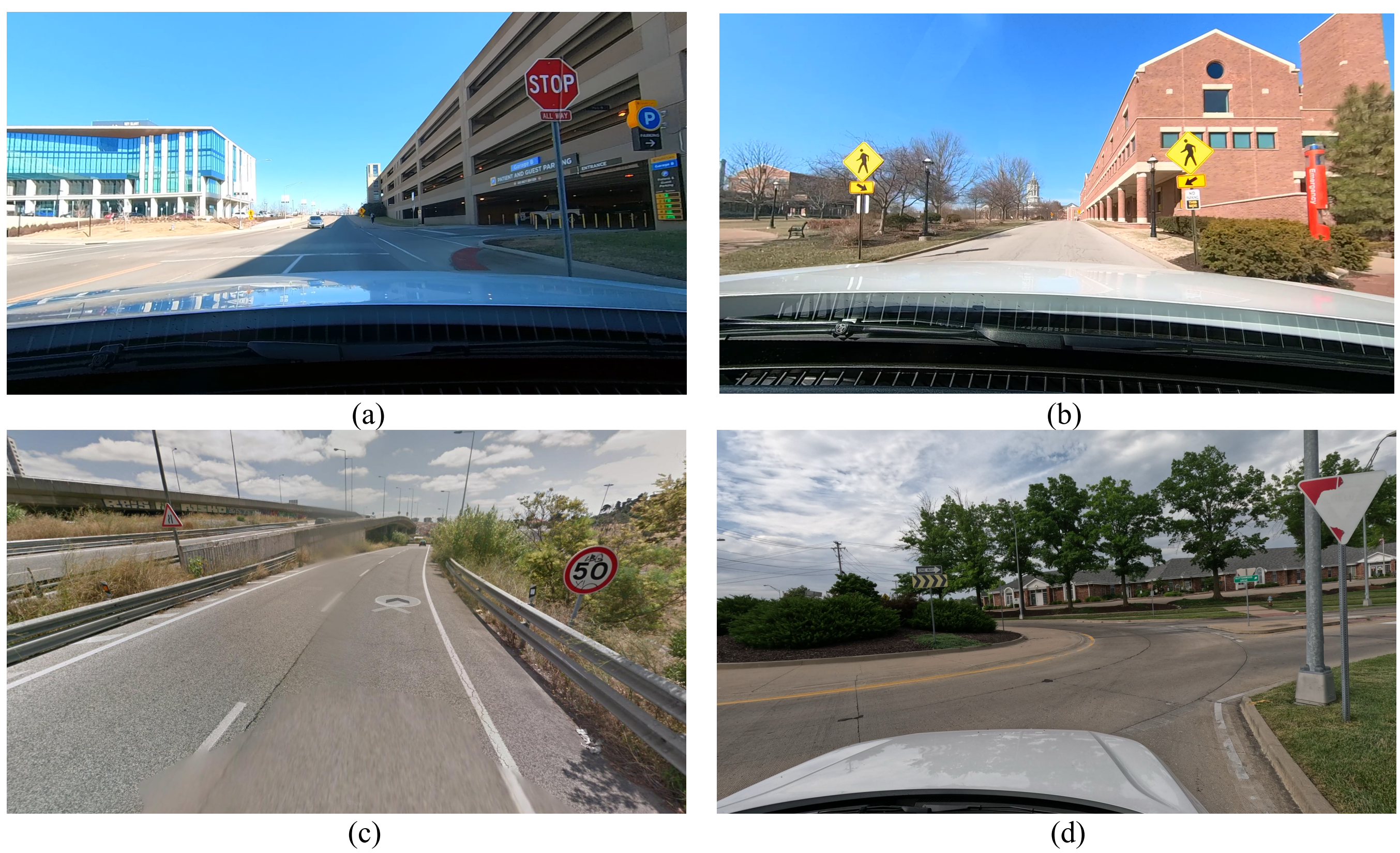

**Figure 3. Traffic Sign Images from Different Source, (a) Augmentation Effect Bent on Normal Sign Image, (b) Sign without Visual Damage, (c) Sign Image from Public Dataset- Damaged Signs Dataset, (d) Damaged Sign Image Manually Collected on Urban Area in Columbia, Missouri**

*Data Fusion and Calibration*

Effective integration of visual and LiDAR data required precise temporal and spatial alignment between the two sensor systems. Temporal synchronization was achieved through ROS timestamp coordination, while spatial alignment utilized 32 randomly distributed correspondence points across the 3D point cloud and 2D image space to compute the extrinsic align point cloud data with image and extract LiDAR measurements. The data fusion process is demonstrated in **Figure 4**. The left images show the scene captured from the GoPro camera and LiDAR respectively, while the right image shows the fusion result utilizing the computed extrinsic transformation matrix. This fused dataset allows us to extract both visual characteristics for daytime assessment and retro-intensity measurements for nighttime evaluation from the same traffic sign instances.

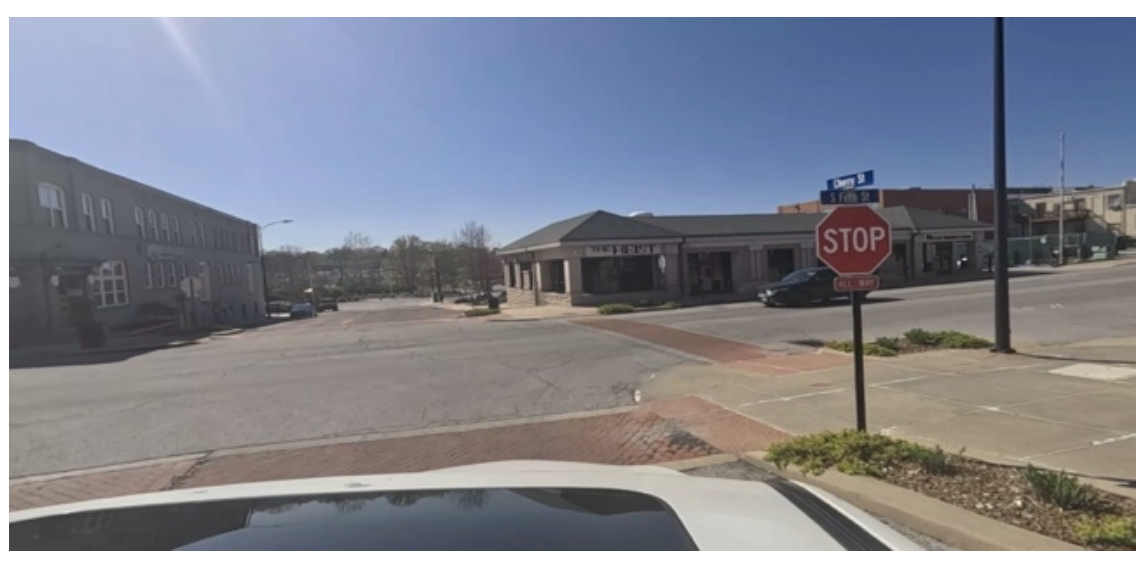

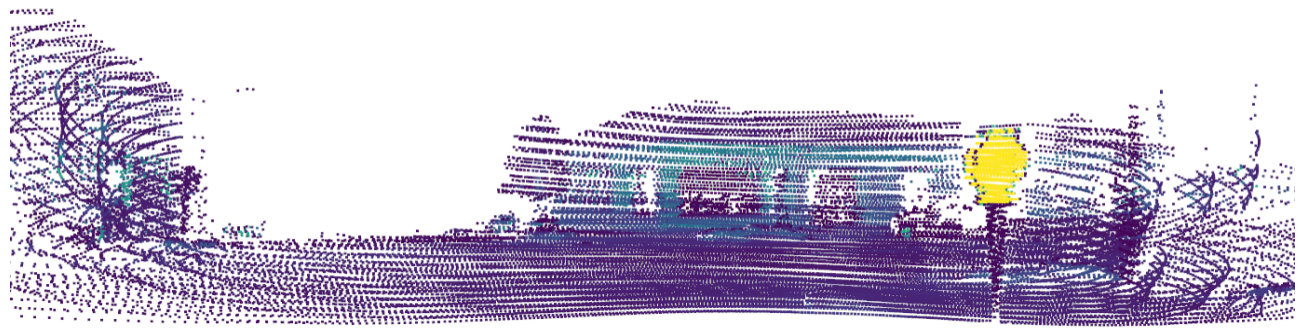

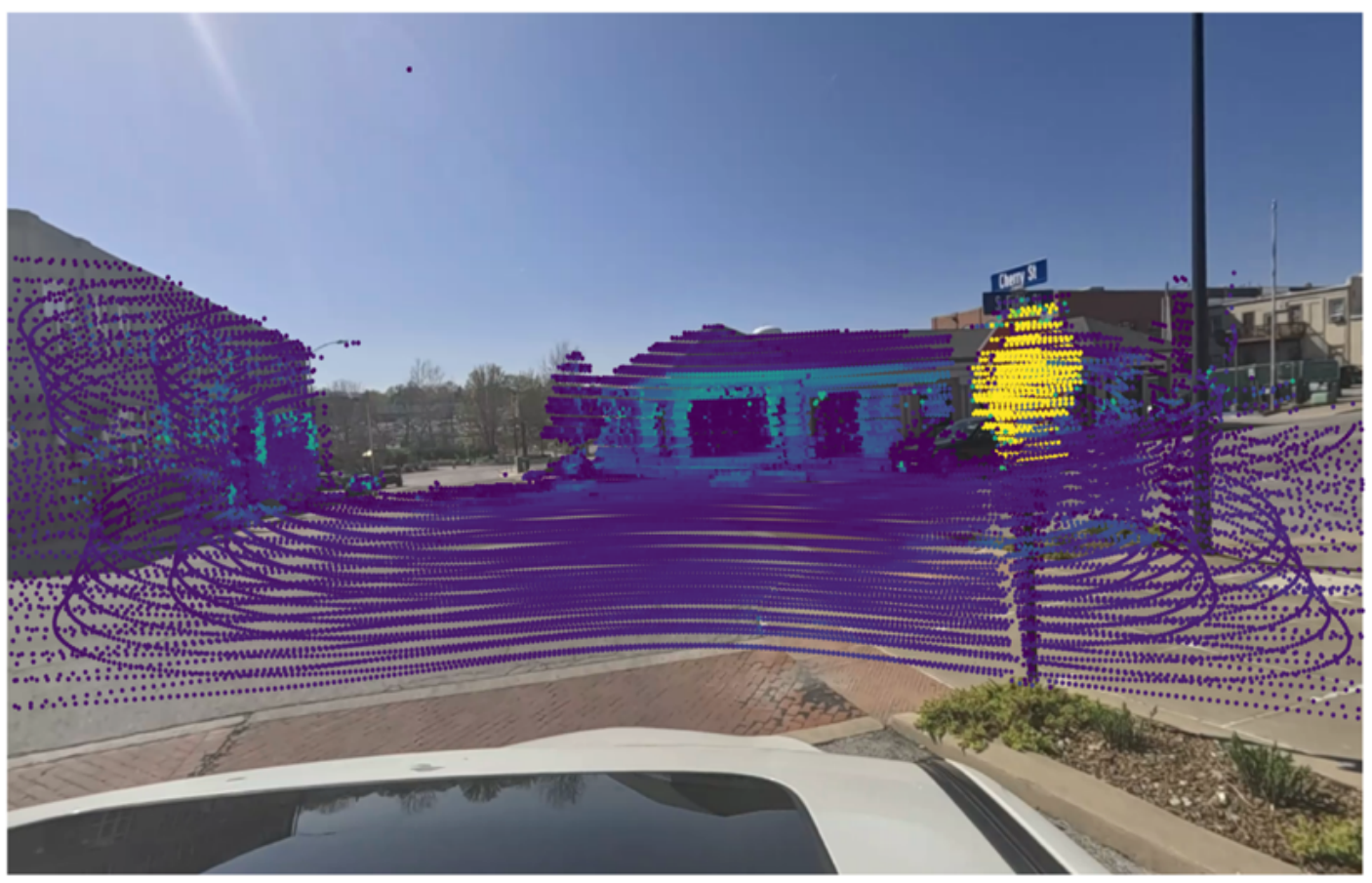

**Figure 4. Data Fusion Process**

**Dual-Mode Evaluation Framework for Traffic Sign Performance Attributes**

The dual-mode assessment framework integrates daytime visual performance evaluation with nighttime retroreflectivity assessment to comprehensively evaluate traffic sign functionality. This framework ensures that signs maintain their essential capability in guiding drivers, pedestrians, and cyclists under both daytime and nighttime conditions through systematic evaluation of complementary performance attributes.

*Daytime Visual Performance Assessment*

The daytime assessment uses fine-tuned VLMs to generate descriptive evaluation of traffic signs across four key factors: legibility, color fidelity, surface and shape integrity, and surrounding environment condition from visual imagery. Each assessment generates five descriptive terms per factor. Three leading VLMs, Qwen2.5-VL (*25*), InternVL (*26*), and LLaVA-NeXT-Video (*27*) were fine-tuned for automated traffic sign inspection using a unified dataset and training configuration.

Qwen2.5-VL is a transformer-based multimodal large language model (MLLM) extended from Qwen2.5 with a Vision Transformer (ViT) encoder and an autoregressive language decoder. To encode visual information, the image input $I \in \mathbb{R}^{HXWX3}$ is divided into patches and embedded into a sequence of visual tokens $V = \{v_1, v_2, \ldots, v_n\} \in \mathbb{R}^{nxd}$, where d is the feature dimension and n is the number of patches. This can be denoted as $V = ViT(I)$. This encoding preserves spatial relationships and semantic features of the traffic sign and its surrounding context, allowing the model to capture nuanced information such as shape integrity or occlusions. The Qwen2.5 decoder is an autoregressive transformer that conditions on both visual tokens $V$ and textual inputs $X$ to generate each output token $y_t$, following $P(y_t|y_{<t}, V, X) = Decoder(y_{<t}, V, X)$. The decoder enables open-ended reasoning for describing physical damage, recognizing symbolic meaning, and assessing visibility compliance.

InternVL is designed for vision-language understanding and generation, with a tightly integrated architecture that supports fine-grained visual reasoning. Unlike models relying on external image encoders (e.g., CLIP or BLIP), InternVL optimizes both visual and language components jointly for a stronger image-text alignment and object-level grounding. It leverages a dual-stream encoder with a Swin Transformer for high-resolution feature extraction, and a visual grounding module to align image regions with language tokens. Given an input image $I \in \mathbb{R}^{HXWX3}$, the visual content is encoded as a token sequence $V \in Swin(I) \in \mathbb{R}^{n\,x\,d}$ , where n is the number of patches and d the embedding dimension. Additionally, object-level features $O = \{o_{1\ldots,} o_k\}$ are extracted via grounding-aware modules for enhanced spatial reasoning. The decoder is an autoregressive transformer that generates responses conditioned on both $V$ and prompt tokens $X$, as expressed by $P(y_t|y_{<t}, V, X) = Decoder(y_{<t}, V, X)$. InternVL's tight integration supports dense reasoning tasks, such as identifying the legibility on signs or determining visibility in complex environments.

LLaVA-NeXT-Video extends the LLaVA-NeXT image-text foundation with a temporal vision encoder for processing video inputs. Its architecture supports frame-level understanding, event-phase alignment, and open-ended generation, making it suitable for traffic sign evaluation. A sequence of frames $\{I_1, I_2, \ldots, I_T\}$, each of dimension H×W×3, is encoded via a shared CLIP-style or ViT-based backbone, producing per-frame embeddings $V_t =$ $ViT(I_t), t = 1, \ldots, T$. To capture temporal dynamics, the frame embeddings are passed into a Temporal Transformer, yielding a fused representation $V_{video} \in \mathbb{R}^{T\,x\,d}$, $V_{video} = TemporalTransformer(\{V_1, V_T\})$. This structure allows the model to track changes in sign visibility, lighting, occlusion, and condition over time. The language decoder, an autoregressive transformer, generate response conditioned on both textual input X and the fused video representation $V_{video}$. It produces token-by-token outputs that reflect temporally-informed visual reasoning: $P(y_t | y_{<t}, V_{video}, X) = Decoder(y_{<t}, V_{video}, X)$. This enables frame-aware multimodal tasks such as motion-consistent object detection, reflectivity assessment, or transient occlusion analysis.

The fine-tuning process incorporated prompt-based supervision, region-specific cropping, and LoRA-based optimization to enhance traffic sign assessment. Multiple instructional prompts were constructed per sample to guide the model in addressing distinct aspects of traffic sign inspection including surface damage identification, text legibility assessment, discoloration detection, and visibility obstruction evaluation. A preprocessing pipeline automatically cropped traffic sign regions to reduce background noise and emphasize relevant features. LoRA adapters were inserted into key transformer layers while freezing the base model, significantly reducing memory footprint and computational cost.

Training was conducted using a batch size of three per GPU with gradient accumulation of eight steps, maximum token length of 4800, and mixed-precision (FP16) training for four epochs using two NVIDIA RTX A6000 GPUs.

*Nighttime Retroreflectivity Assessment*

The nighttime retroreflectivity assessment operates in parallel with daytime assessment by analyzing LiDAR point cloud data within regions of interest (ROI) identified by YOLOv8. Since cropped images often include both target signs and surrounding features, we implemented Otsu thresholding based on distance values to establish binary segmentation between foreground sign regions and background areas to address this spatial disambiguation challenge. By leveraging spatial correspondence between point cloud and image, this approach extracts retro-intensity values directly from traffic sign surfaces while removing background noise.

However, laser-based retro-intensity measurements exhibit inherent variability due to beam distance (*28*) and incidence angle effects (*11*, *29*). Without proper correction, these geometric dependencies introduce significant errors in traffic sign retroreflectivity evaluation. To address these biases, LiDAR calibration is essential for accurate measurements.

Following the empirical calibration methodology established by Voegtle and Wakaluk (*30*) and Ai's previous work (*11*). Joshua et al. (*31*) developed a comprehensive calibration framework for retro-intensity normalization. Their approach models the retro-intensity$\rho(\theta, D)$ as the product of two distinct functions: $f(\theta)$, which captures angular dependence, and $g(D)$, which accounts for distance-related intensity variations. Drawing on established photometric principles (*32*, *33*), the angular correction function incorporates classical light illumination models using an empirical Phong surface formulation.

For retroreflective materials, the angular correction function is expressed as:

$$f(\theta) = \left[\left(1 - k_s(\theta)\right)\cos(\theta) + k_s(\theta)\right] \tag{1}$$

where $k_s\ (\theta)$ represents a second-order polynomial function calibrated for Type I engineer grade signs, and the distance correction follows $g(D) = D\hat{}\alpha$. The complete normalization model integrates these components:

$$\rho(\vartheta, D) = \left[\left(1 - k_s(\theta)\right)\cos(\theta) + k_s(\theta)\right] \cdot D^{\alpha} \tag{2}$$

The normalization process applies these correction factors to eliminate geometric effects from raw measurements. The normalized retro-intensity is calculated as:

$$\rho_{norm}(\vartheta, D) = \frac{\rho_{raw}(\vartheta, D)}{\left[\left(1 - k_s(\theta)\right)\cos(\theta) + k_s(\theta)\right] \cdot D^{\alpha}} \tag{3}$$

where $\rho_raw\ (\vartheta, D)$ represents the measured retro-intensity.

Given that this study utilizes the same LiDAR sensor model as Joshua et al. (*31*), we directly adopted their calibration parameters for retro-intensity normalization. Following Ai and Tsai's (*11*) study, the estimated retroreflectivity is defined as:

$$R_{est} = -285.9 + 392.3\,\rho_{norm}(\vartheta, D) \tag{4}$$

**Sign Condition Index Integration**

The dual-mode traffic sign evaluation result requires a unified scoring system to convert individual assessment components into quantitative condition metrics. To address this need, we developed the SCI, a systematic scoring framework that integrates visual performance assessments from the fine-tuned VLMs with LiDAR-derived retroreflectivity estimation.

*Hybrid Scoring System for Daytime Assessment*

While the VLMs produce five descriptive words per visual factor, these qualitative outputs are not directly feasible for maintenance decision making. Numerical conversion is necessary to bridge the gap between textual descriptions and quantitative metrics. Therefore, a hybrid scoring system was developed that leverages both sentiment and multimodal alignment techniques.

Sentiment analysis, a natural language processing technique that determines the emotional polarity, was employed to evaluate whether the descriptive words from the fine-tuned VLMs indicate positive or negative traffic sign conditions. CLIP represents a multimodal model that facilitates assessment of alignment between textual descriptions and corresponding visual content. Each set of five descriptive words undergoes dual evaluation through these complementary methods. The sentiment analysis component assigns valence scores to traffic sign condition related terminology, while CLIP scoring measures the consistency between generated descriptions and actual visual characteristics from the image.

The hybrid method employs a weighted integration approach, combining 70% sentiment scores with 30% CLIP alignment scores to generate composite evaluation metrics. This weighting was selected as a heuristic to emphasize condition-related semantic polarity while retaining image-text consistency through CLIP alignment. The resulting composite scores are normalized into continuous numerical scores for systematic condition assessment.

*Retroreflectivity Scoring Rules*

The nighttime evaluation component focuses exclusively on retroreflectivity assessment, utilizing the established linear relationship between normalized LiDAR retro-intensity and actual retroreflectivity measurements. Following the methodology developed by Ai and Tsai (*11*), the linear regression provides the foundation for retroreflectivity estimation is shown in **Equation 4**.

However, the previous linear model was calibrated and normalized specifically using Type I Stop Sign, which show restriction to other sign types. According to MUTCD minimum retroreflectivity requirements which is shown in **Figure 5**, yellow and orange background signs cannot be manufactured using Type I materials and must utilize Type II or higher-grade materials. For these sign categories, direct application of the **Equation 4** may not provide accurate retroreflectivity estimates.

**Table 2A-5. Minimum Maintained Retroreflectivity Levels[1]**

| Sign Color | Beaded Sheeting Type (ASTM D4956) | | | Prismatic Sheeting | Additional Criteria |
|---|---|---|---|---|---|
| | I | II | III | | |
| White on Green | W*; G ≥ 7 | W*; G ≥ 15 | W*; G ≥ 25 | W ≥ 250; G ≥ 25 | Overhead |
| | W*; G ≥ 7 | W ≥ 120; G ≥ 15 | | | Post-mounted |
| White on Blue | W*; B ≥ 3 | W*; B ≥ 5 | W*; B ≥ 12 | W ≥ 250; B ≥ 12 | Overhead |
| | W*; B ≥ 3 | W ≥ 120; B ≥ 7 | | | Post-mounted |
| White on Brown | W*; Br ≥ 1 | W*; Br ≥ 5 | W*; Br ≥ 10 | W ≥ 350; Br ≥ 10 | Overhead |
| | W*; Br ≥ 1 | W ≥ 150; Br ≥ 5 | | | Post-mounted |
| Black on Yellow or Black on Orange | Y*; O* | Y ≥ 50; O ≥ 50 | | | 2 |
| | Y*; O* | Y ≥ 75; O ≥ 75 | | | 3 |
| White on Red | W ≥ 35; R ≥ 7 | | | | 4 |
| Black on White | W ≥ 50 | | | | – |

[1] The minimum maintained retroreflectivity levels shown in this table are in units of $cd/lx/m^2$ measured at an observation angle of 0.2° and an entrance angle of -4.0°.
[2] For word legend and fine symbol signs measuring at least 48 inches and for all sizes of bold symbol signs
[3] For word legend and fine symbol signs measuring less than 48 inches
[4] Minimum sign contrast ratio ≥ 3:1 (white retroreflectivity ÷ red retroreflectivity)
* This sheeting type shall not be used for this color for this application

**Bold Symbol Signs**

- W1-1,2 – Turn and Curve
- W1-3,4 – Reverse Turn and Curve
- W1-5 – Winding Road
- W1-6,7 – Large Arrow
- W1-8 – Chevron
- W1-10 – Intersection in Curve
- W1-11 – Hairpin Curve
- W1-15 – 270 Degree Loop
- W2-1 – Cross Road
- W2-2,3 – Side Road
- W2-4,5 – T and Y Intersection
- W2-6 – Circular Intersection
- W2-7,8 – Double Side Roads
- W3-1 – Stop Ahead
- W3-2 – Yield Ahead
- W3-3 – Signal Ahead
- W4-1 – Merge
- W4-2 – Lane Ends
- W4-3 – Added Lane
- W4-5 – Entering Roadway Merge
- W4-6 – Entering Roadway Added Lane
- W6-1,2 – Divided Highway Begins and Ends
- W6-3 – Two-Way Traffic
- W10-1,2,3,4,11,12 – Grade Crossing Advance Warning
- W11-2 – Pedestrian Crossing
- W11-3,4,16-22 – Large Animals
- W11-5 – Farm Equipment
- W11-6 – Snowmobile Crossing
- W11-7 – Equestrian Crossing
- W11-8 – Fire Station
- W11-10 – Truck Crossing
- W12-1 – Double Arrow
- W16-5P,6P,7P – Pointing Arrow Plaques
- W20-7 – Flagger
- W21-1 – Worker

**Fine Symbol Signs** (symbol signs not listed as bold symbol signs)

**Special Cases**

- W3-1 – Stop Ahead: Red retroreflectivity ≥ 7
- W3-2 – Yield Ahead: Red retroreflectivity ≥ 7; White retroreflectivity ≥ 35
- W3-3 – Signal Ahead: Red retroreflectivity ≥ 7; Green retroreflectivity ≥ 7
- W3-5 – Speed Reduction: White retroreflectivity ≥ 50
- For non-diamond shaped signs, such as W14-3 (No Passing Zone), W4-4P (Cross Traffic Does Not Stop), or W13-1P,2,3,6,7 (Speed Advisory Signs), use the largest sign dimension to determine the proper minimum retroreflectivity level.

**Figure 5. Minimum Maintained Retroreflectivity Levels from MUTCD**

**Table 2. Minimum Retro-Intensity Values for Acceptable Condition for Various Colors and Sheeting Types of Traffic Signs**

| Red, EG | Red, MP | White, EG | White, MP | Yellow, MP | Green, EG | Green, MP | Blue, EG | Blue, MP |
|---|---|---|---|---|---|---|---|---|
| 0.705 | 0.780 | 0.735 | 0.750 | 0.762 | 0.700 | 0.740 | 0.740 | 0.755 |

To address this limitation, we adopted the categorical assessment thresholds established by Steele et al. (*9*), who developed retro-intensity ranges for various sign colors and materials through comprehensive field validation. Table 2 represents the detailed threshold that Steele et al. developed. Given this standard above, the retroreflectivity scoring uses a three-tier rule: signs meeting MUTCD or rules in **Table 2**, receive 5 points, those at 60-99% of standards receive 3 points, and those below 60% receive 1 point. This simplified approach provides clear maintenance guidance while maintaining consistency with the daytime evaluation framework.

*Final Score Integration*

Considering the functional importance of retroreflectivity and legibility, poor performance in either area will lead to failure in delivering transportation information to road users. Therefore, if the legibility score is below 1 or the retroreflectivity score equals 1, the final SCI is automatically assigned a value of 1, indicating immediate replacement needs. For all other cases, the final SCI is calculated using a weighted scoring scheme showing in Equation (5).

$$SCI = 60\% Legibility + 40\% Color + 20\% Surface\ and\ Shape\ Integrity + 20\% Surrounding\ Envorinment\ Codition + 60\% Retroreflectivity \quad (5)$$

Equation (5) is applied only when the legibility score is greater than or equal to 1 and the retroreflectivity score is greater than 1. The resulting SCI provides maintenance guidance where scores range from 0 to 10. With test using extreme words, the score between 1 to 5 indicate replacement needs, from 5 to 7 suggests monitoring, and from 7 to 10 recommend continued use with routine maintenance.

## RESULT AND DISCUSSION

The proposed framework was evaluated through comprehensive testing of both individual components and integrated system performance. A test set of 412 images was used for VLM assessment, while a validation set of 462 traffic signs with corresponding LiDAR data evaluated the complete dual-mode framework and SCI integration.

### Dual-Mode Assessment Results

*Daytime Visual Performance Assessment Results*

Three VLMs (Qwen2.5-VL, InternVL, and LLaVA-NeXT-Video) were evaluated using bidirectional cosine similarity scores to measure semantic consistency between generated descriptive terms and ground truth annotations across four assessment factors.

Cosine Similarity Prediction to Ground Truth: $S_{P_G} = \frac{1}{m}\sum_{i=1}^{m} max_{j=1}^{n} \cos\left(Emb(p_i), Emb(g_j)\right)$ (6)

Cosine Similarity Ground Truth to Prediction: $S_{G_P} = \frac{1}{n}\sum_{j=1}^{n} max_{i=1}^{m} \cos\left(Emb(g_j), Emb(p_i)\right)$ (7)

Bidirectional Cosine Similarity Score: $S = \frac{S_{P_G}+S_{G_P}}{2}$ (8)

Where $m$ and $n$ denotes the number of words in predicted and ground truth sets. The bidirectional cosine similarity ranges from 0 to 1, a higher value indicates a more accurate semantic alignment. **Figure 5** and **Figure 6** present the cosine similarity score of each factor and the overall score of Qwen2.5VL, LLaVA-NeXT-Video, InternVL models. **Table 3** shows the statistics of bidirectional cosine similarity score of VLMs.

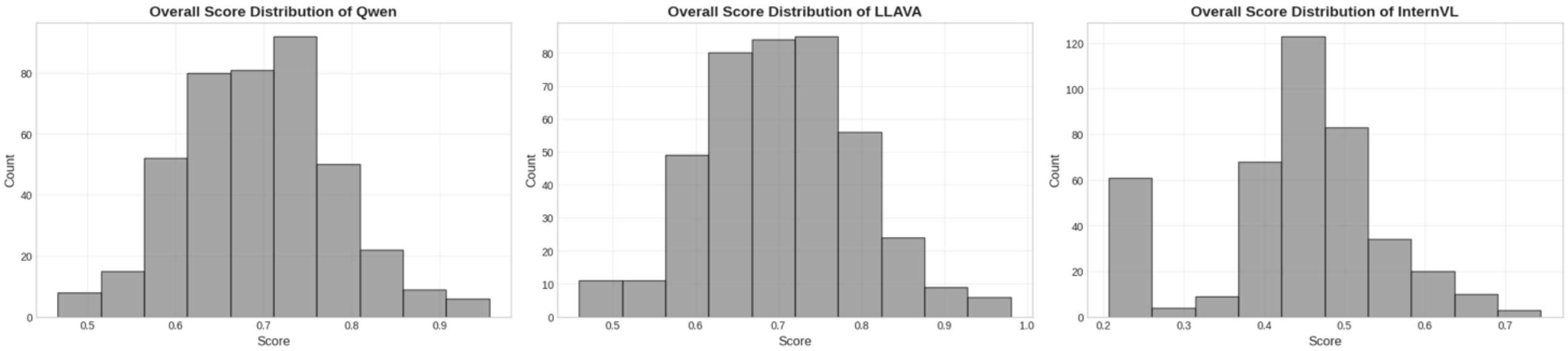

**Figure 5. Overall Cosine Similarity Score of Qwen, LLaVA, and InternVL.**

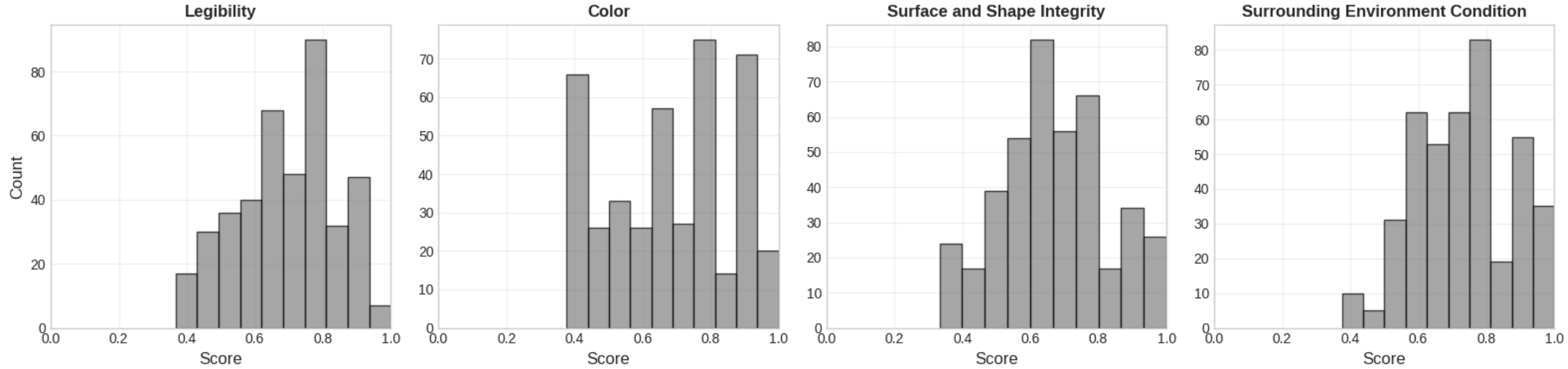

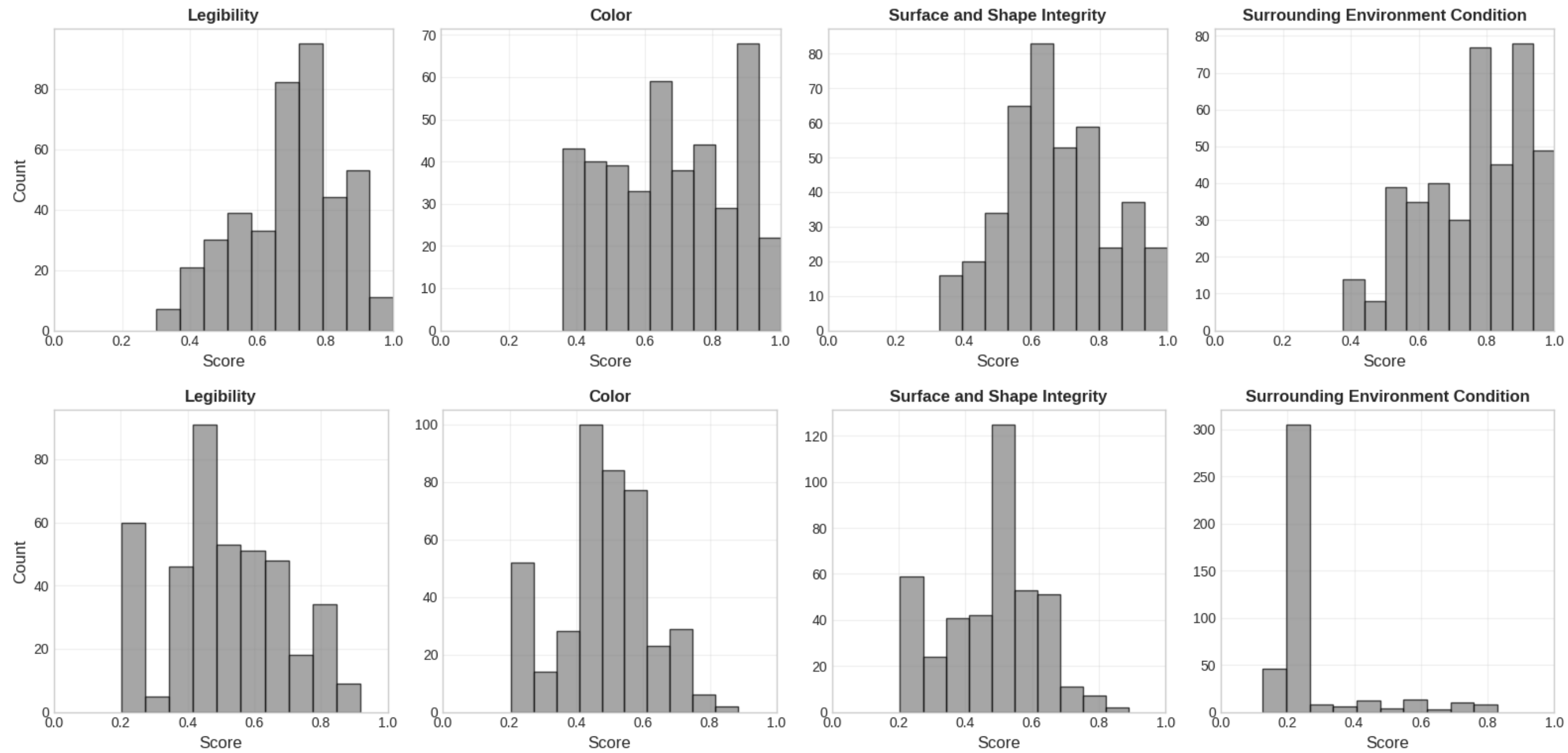

**Figure 6. Cosine Similarity Score of Each Factor for Qwen (up), LLaVA (middle), and InternVL(bottom)**

**Table 3. Bidirectional Cosine Similarity Performance of VLMs across Assessment Factors**

| VLM | Legibility | | | Color | | | Surface and Shape Integrity | | | Surround Environment Condition | | |
|---|---|---|---|---|---|---|---|---|---|---|---|---|
| | Mean | Median | S.D | Mean | Median | S.D | Mean | Median | S.D | Mean | Median | S.D |
| Qwen | 0.691 | 0.693 | 0.143 | 0.683 | 0.688 | 0.182 | 0.671 | 0.661 | 0.160 | 0.734 | 0.720 | 0.144 |
| LLAVA | 0.700 | 0.714 | 0.148 | 0.677 | 0.674 | 0.180 | 0.673 | 0.660 | 0.157 | 0.763 | 0.790 | 0.159 |
| IntenVL | 0.516 | 0.498 | 0.143 | 0.482 | 0.490 | 0.137 | 0.489 | 0.511 | 0.144 | 0.282 | 0.247 | 0.134 |

Based on **Table 3**, **Figure 5** and **Figure 6**, LLaVA achieves the highest performance across all factors with mean scores around 0.673-0.763, followed by Qwen with means score around 0.671-0.734, while InternVL lagged behind around 0.282-0.516. All models perform best on Surrounding Environment Condition, with LLaVA and Qwen exceeding 0.73. The score distributions reveal that LLaVA and Qwen exhibit consistent high-quality predictions with peaks around 0.6-0.8, while InternVL demonstrates greater variance and lower center tendencies.

*Nighttime Retroreflectivity Assessment Results*

Following the Ai and Tsai, and Joshua's research, LiDAR retro-intensity calibration was applied to 1,078 samples across 14 distinct sign types. As shown in **Figure 7**, raw and normalized intensities exhibit strong positive correlation, validating calibration effectiveness. Notably, some signs with high raw retro-intensity values show substantial reduction after normalization. This emphasizes the significant influence from certain angles and distances.

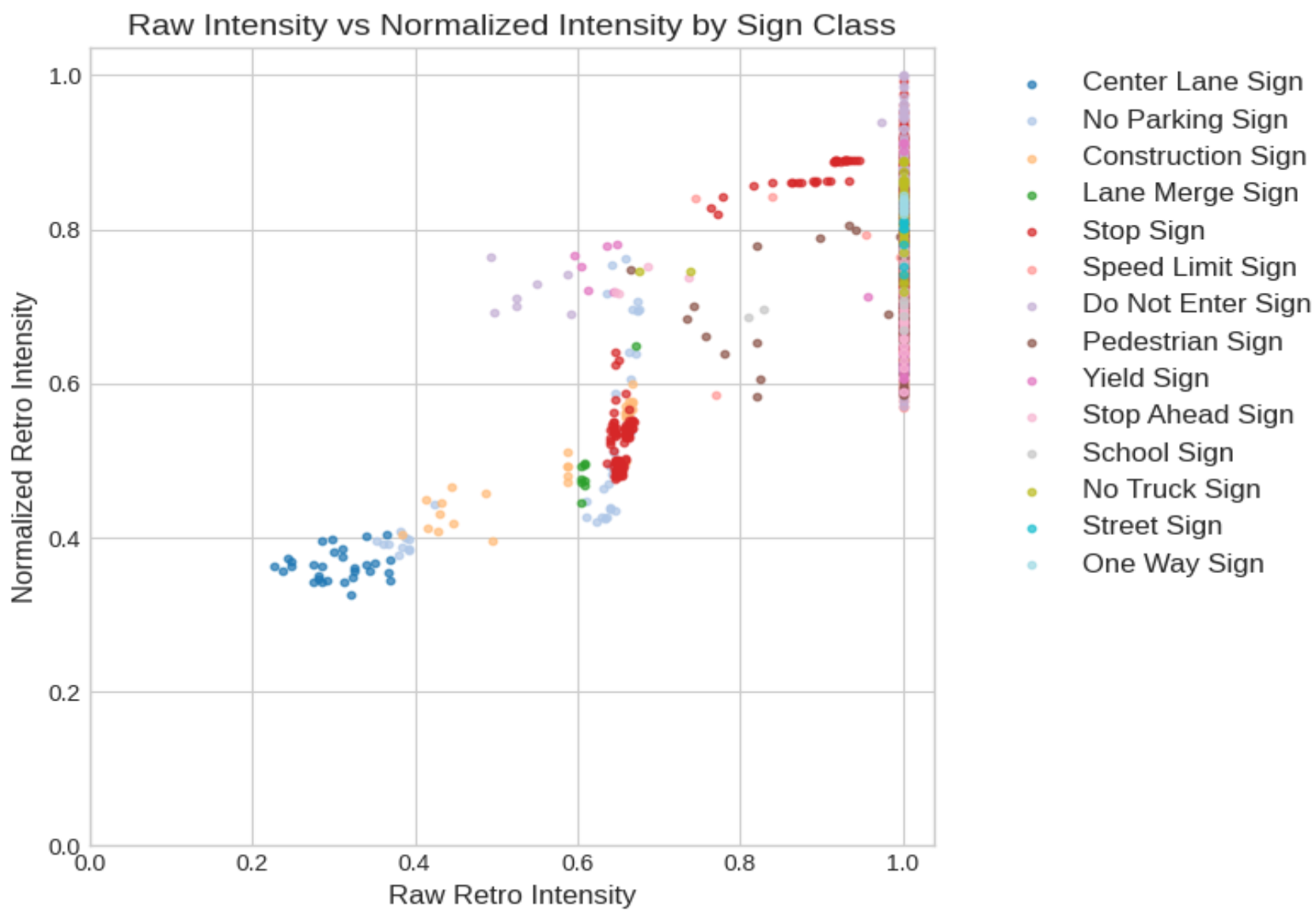

**Figure 7. Normalized Retro-Intensity VS Raw Retro-Intensity**

**SCI Integration Result**

The SCI integrates daytime and nighttime assessment into a unified condition score. The hybrid scoring approach, using 70% sentiment analysis and 30% CLIP scoring, converted VLM-generated descriptive terms into continuous numerical scores for each daytime assessment factor. For nighttime assessment, a straightforward scoring system was developed based on the thresholds established by Ai and Tsai (*11*) and Steele et al. (*9*). Given the significant importance of legibility and retroreflectivity for traffic sign functionality, either factor receiving a score of 1 automatically assigns the final SCI a value of 1. **Table 4** illustrates the daytime visual performance scoring results using representative test set examples. Without considering retroreflectivity, the daytime assessment showed scores of 3, 2, 1, and 1 for Legibility, Color, Surface and Shape Integrity, and Surrounding Environment Condition respectively, totaling 7 points for daytime performance. The descriptive terms in each column are predicted using LLaVA.

**Table 4. Daytime Visual Performance Scoring Results Using Representative Test Set**

| Image | Source and Overall Score | Legibility | Color | Surface and Shape Integrity | Surrounding Environment Condition |
|---|---|---|---|---|---|
| | Damaged Signs Dataset | Fuzzy, Blurred, Unclear, Indistinct, Distorted | Faded, Dull, Washed-out, Muted, Subdued | Intact, Unbroken, Solid, Complete, Undamaged | Unobstructed, Clear, Open, Uncluttered, Visible |
| | 2.712 | 0.859 | 0.949 | 0.449 | 0.454 |
| | Augmentation | Clear, Distorted, Partially Obstructed, Partially Blurred, Partially Unclear | Faded, Dull, Washed Out, Pale | Intact, Undamaged, Unwarped, Straight, Symmetrical | Unobstructed, Clear, Open, Visible, Uncluttered |
| | 2.890 | 0.854 | 0.890 | 0.598 | 0.548 |

| | Damaged Signs Dataset | Fuzzy, Blurred, Unclear, Indistinct, Distorted | Faded, Dull, Washed, Muted | Intact, Unbroken, Solid, Complete, Whole | Unobstructed, Clear, Open, Spacious, Uncluttered |
|---|---|---|---|---|---|
| | 2.997 | 0.945 | 0.940 | 0.596 | 0.516 |
| | Manually Collected | Clear, concise, easy to read, well-defined, precise | Vibrant, accurate, visible, consistent, attention-grabbing | Intact, sturdy, well-constructed, durable, secure | Unobstructed, clear, open, free from clutter, and easily visible |
| | 5.185 | 1.969 | 1.463 | 0.755 | 0.998 |
| | Manually Collected | Clear, concise, easy to read, well-defined, precise | Vibrant, accurate, visible, consistent, attention-grabbing | Intact, sturdy, well-constructed, secure, reliable | Unobstructed, clear, open, free from clutter, easily visible |
| | 5.183 | 1.977 | 1.465 | 0.745 | 0.996 |
| | Manually Collected | Clear, Distinct, Easy to Read, Well-defined, Precise | Vibrant, Accurate, Visible, Saturated, Consistent | Intact, Unbroken, Solid, Stable, Secure | Unobstructed, Clear, Open, Accessible, Unhindered |
| | 5.142 | 2.128 | 1.462 | 0.750 | 0.802 |
| | Manually Collected | Clear, Distinct, Easy to Read, Well-defined, Precise | Vibrant, Saturated, Accurate, Visible, Consistent | Intact, Undamaged, Structurally Sound, Symmetrical, Well-maintained | Unobstructed, Clear View, Free from Obstacles, Safe for Pedestrians, Well-positioned |
| | 5.381 | 2.146 | 1.468 | 0.775 | 0.992 |

After filtering unqualified samples in validation set, 462 traffic signs received final SCI scores. Among these, 68 of the signs were assigned an SCI of 1. All these cases are caused by inadequate retroreflectivity performance. **Figure 8** presents examples of signs requiring immediate replacement, where the normalized retro-intensity values from left to right are 0.326, 0.534, and 0.539, respectively. For these stop signs, the estimated retroreflectivity values are all negative, which are meaningless and fails to meet MUTCD minimum standards. The detailed descriptive terms and corresponding hybrid scores for **Figure 8** are presented in **Table 5**.

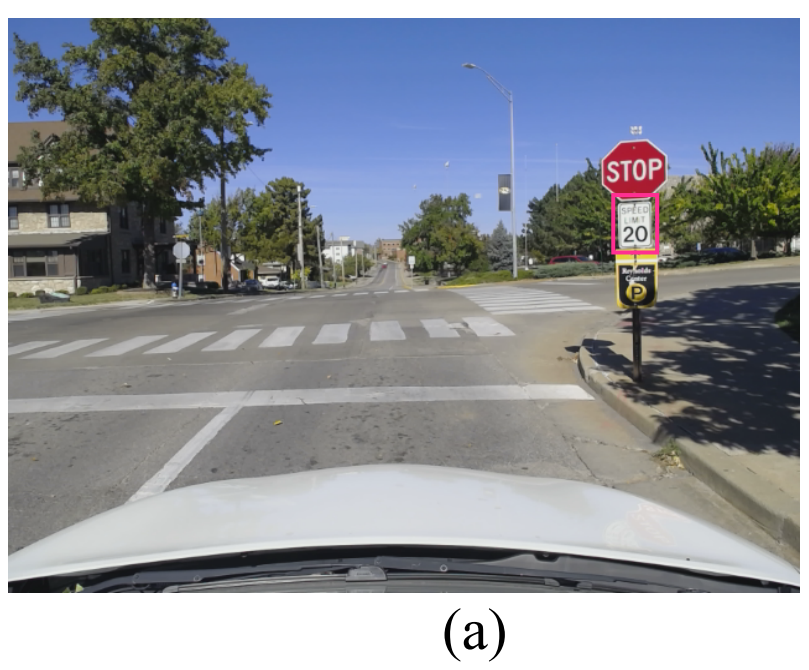

(a)

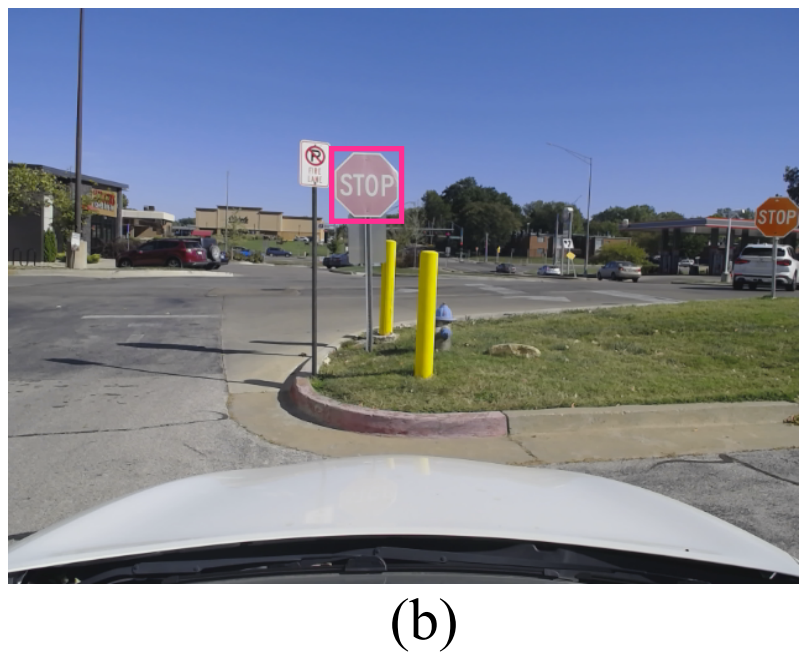

(b)

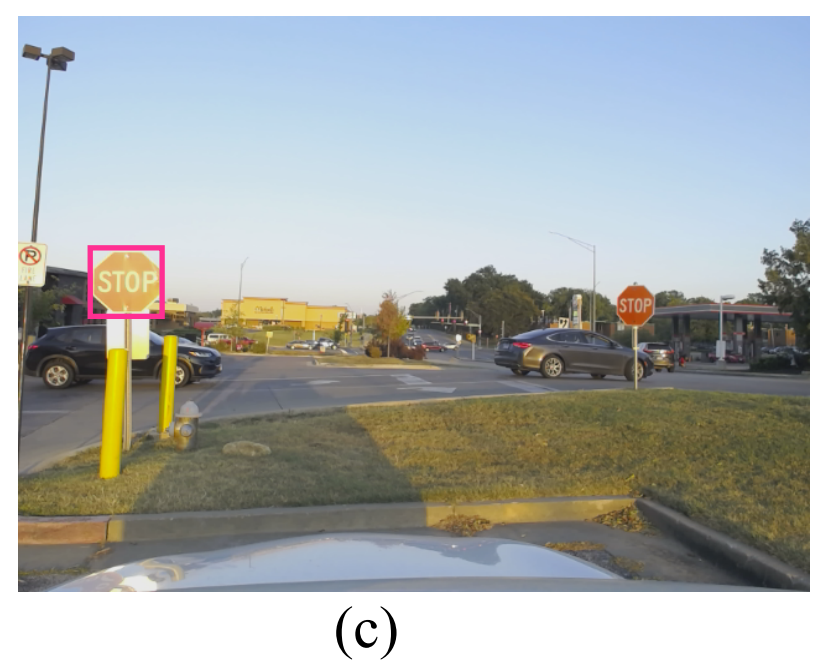

(c)

**Figure 8. SCI of 1 Due to Low Retroreflectivity**

**Table 5. SCI Factor-wise Assessment Result**

| Image | Overall Score | Legibility | Color | Surface and Shape Integrity | Surrounding Environment Condition | Retroreflectivity |
|---|---|---|---|---|---|---|
| | 1 | Clear, Distinct, Easy to read, Well-defined, Precise | Faded, Dull, Washed Out, Pale, Subdued | Intact, Unbroken, Symmetrical, Well-maintained, Solid | Unobstructed, Clear, Open, Uncluttered, Visible | Normalized Value: 0.326 |
| | | 2.104 | 0.957 | 0.514 | 0.580 | 1 |
| | 1 | Clear, Distorted, Partially Obstructed, Partially Blurred, Partially Unclear | Faded, Dull, Washed Out, Pale | Intact, Undamaged, Unwarped, Straight, Symmetrical | Unobstructed, Clear, Open, Visible, Uncluttered | Normalized Value: 0.534 |
| | | 0.854 | 0.890 | 0.598 | 0.548 | 1 |
| | 1 | Fuzzy, Blurred, Unclear, Indistinct, Distorted | Faded, Dull, Washed, Muted | Intact, Unbroken, Solid, Complete, Whole | Unobstructed, Clear, Open, Spacious, Uncluttered | Normalized Value: 0.539 |
| | | 0.945 | 0.940 | 0.596 | 0.516 | 1 |

## CONCLUSION

This study presents a unified framework for traffic sign condition assessment, integrating daytime visual performance assessment using fine-tuned VLMs and nighttime retroreflectivity estimation based on LiDAR-derived retro-intensity. A hybrid scoring system utilizing 70% sentiment analysis and 30% CLIP scoring successfully converted VLM-generated descriptive outputs into standardized numerical scores, which combined with straightforward retroreflectivity scoring rules, establishing the SCI to give comprehensive maintenance decisions.

Evaluation results demonstrated that LLaVA and Qwen outperformed InternVL, with bidirectional cosine similarity scores of 0.67-0.76 across all factors. The LiDAR calibration normalized raw intensity effectively, reducing angle and distance bias. Among 462 validated traffic signs, 68 of them were flagged by the proposed framework as requiring immediate replacement, due to inadequate retroreflectivity performance.

The framework provides significant practical value by offering a cost-effective alternative to traditional manual inspections and expensive retroreflectometer equipment. Unlike previous research focusing on isolated assessment factors, this study presents an integrated system addressing both daytime and nighttime evaluation requirements within a unified framework. The standardized SCI scoring system enables objective maintenance prioritization, reducing subjective evaluation inconsistencies while providing clear guidance for municipal agencies with limited resources.

However, several limitations should be acknowledged for future research consideration. The linear regression model for retro-intensity to retroreflectivity conversion was calibrated exclusively for Type I materials, restricting its application to signs using higher grade materials. While we incorporated alternative assessment thresholds for different sign types, the framework still lacks unified assessment standards for certain sign colors such as orange construction signs. Additionally, the framework does not separate MUTCD requirements for sign backgrounds versus foreground text and symbols, using 75% of the intensity distribution as a representative value introduces potential bias. The hybrid scoring system also introduces sensitivity to semantic variations, where equivalent terms may yield different numerical scores. In addition, the 70/30 weighting between sentiment analysis and CLIP scoring was heuristically selected, and future work should conduct sensitivity analysis to examine how different weighting schemes affect SCI outcomes. The VLMs also does not consistently generate exactly five descriptive words for each factor as required, necessitating mean imputation for missing entries. Future research should focus on developing unified retroreflectivity assessment standards across all sign types and materials, implementing more robust scoring schemes that reduce semantic sensitivity, and incorporating separate assessment protocols for foreground and background elements to align with MUTCD specifications.

**AUTHOR CONTRIBUTIONS**

The authors confirm contribution to the paper as follows: study conception and design: Zhang, Adu-Gyamfi; data collection: Zhang, Yu, Watts; analysis and interpretation of results: Zhang, Owor, Adu-Gyamfi; draft manuscript preparation: Zhang, Owor, Yu, Watts, Adu-Gyamfi. All authors reviewed the results and approved the final version of the manuscript.